\documentclass[10pt,twocolumn,letterpaper]{article}

\usepackage{iccv}
\usepackage{times}
\usepackage{epsfig}
\usepackage{graphicx}
\usepackage{amsmath}
\usepackage{amssymb}

\graphicspath{{images/}}

\usepackage[pagebackref=true,breaklinks=true,letterpaper=true,colorlinks,bookmarks=false]{hyperref}

\iccvfinalcopy 

\def\iccvPaperID{XXX} 

\ificcvfinal\fi
\begin{document}

\title{Semantic Instance Segmentation with a Discriminative Loss Function}

\author{Bert De Brabandere\thanks{Authors contributed equally}\qquad Davy Neven$^*$\qquad Luc Van Gool\\
ESAT-PSI, KU Leuven\\
{\tt\small firstname.lastname@esat.kuleuven.be}
}

\maketitle

\begin{abstract}
Semantic instance segmentation remains a challenging task. In this work we propose to tackle the problem with a discriminative loss function, operating at the pixel level, that encourages a convolutional network to produce a representation of the image that can easily be clustered into instances with a simple post-processing step.
The loss function encourages the network to map each pixel to a point in feature space so that pixels belonging to the same instance lie close together while different instances are separated by a wide margin.
Our approach of combining an off-the-shelf network with a principled loss function inspired by a metric learning objective is conceptually simple and distinct from recent efforts in instance segmentation. 
In contrast to previous works, our method does not rely on object proposals or recurrent mechanisms.
A key contribution of our work is to demonstrate that such a simple setup without bells and whistles is effective and can perform on-par with more complex methods. Moreover, we show that it does not suffer from some of the limitations of the popular detect-and-segment approaches.
We achieve competitive performance on the Cityscapes and CVPPP leaf segmentation benchmarks.
\end{abstract}

\section{Introduction}
\label{sec:introduction}
\begin{figure}[t]
	\begin{center}
		\includegraphics[width=1.0\linewidth]{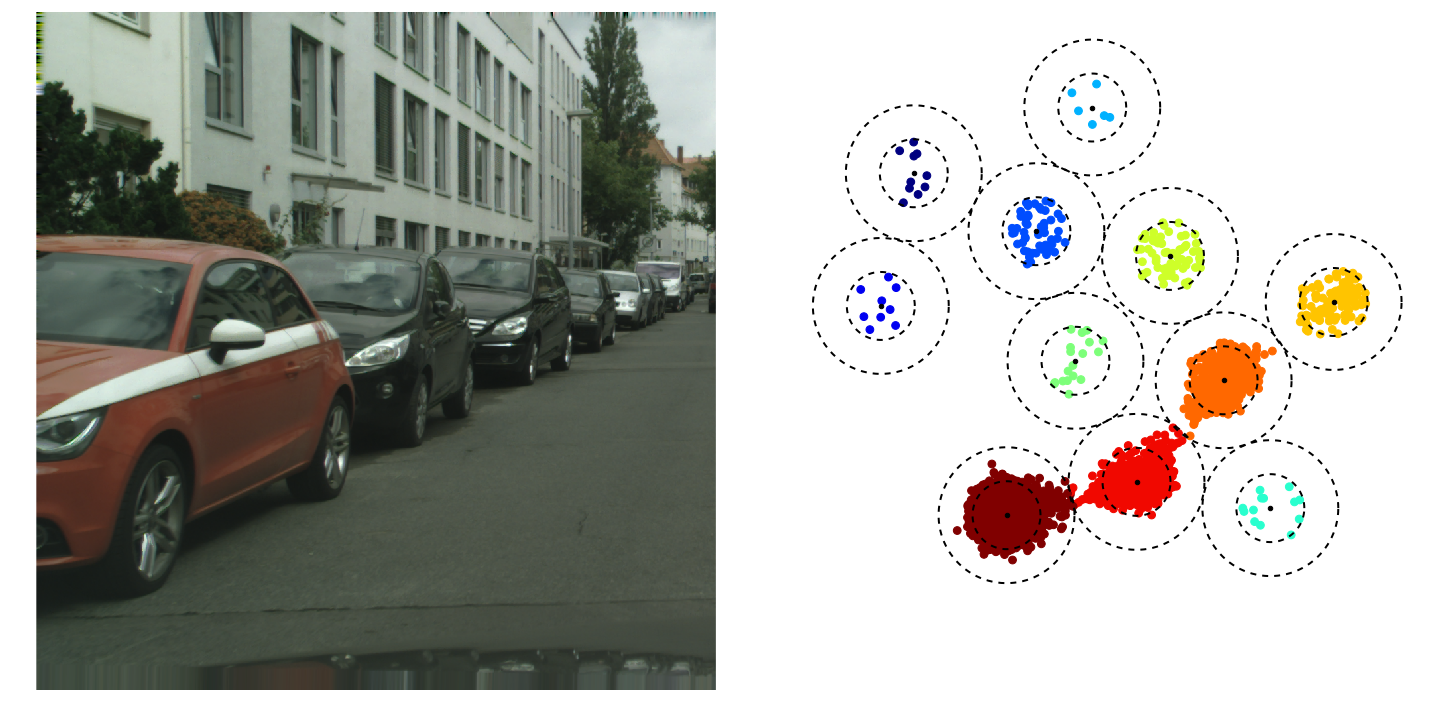}
		\includegraphics[width=1.0\linewidth]{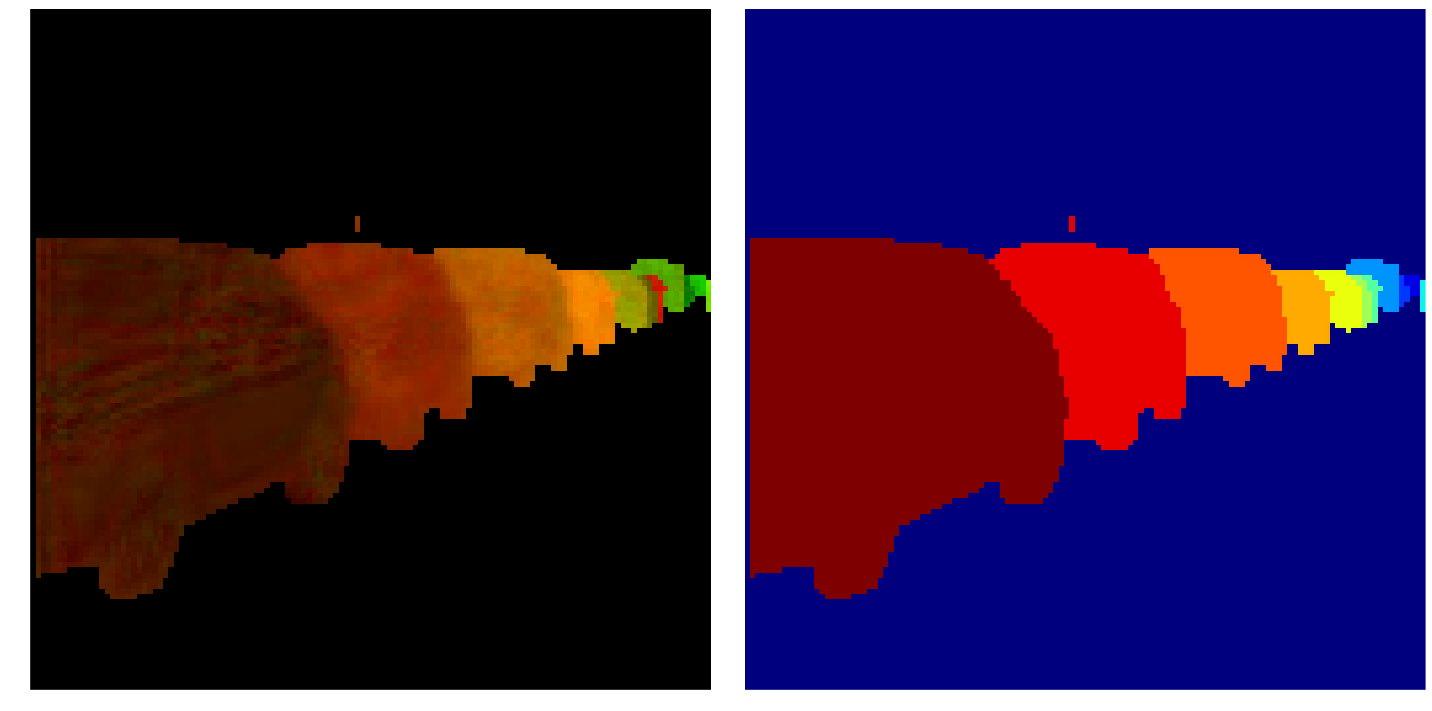}
	\end{center}
	\caption{The network maps each pixel to a point in feature space so that pixels belonging to the same instance are close to each other, and can easily be clustered with a fast post-processing step. From top to bottom, left to right: input image, output of the network, pixel embeddings in 2-dimensional feature space, clustered image.}
	\label{fig:converged_leafs}
\end{figure}

Semantic instance segmentation has recently gained in popularity. As an extension of regular semantic segmentation, the task is to generate a binary segmentation mask for each individual object along with a semantic label. It is considered a fundamentally harder problem than semantic segmentation - where overlapping objects of the same class are segmented as one - and is closely related to the tasks of object counting and object detection. One could also say instance segmentation is a generalization of object detection, with the goal of producing a segmentation mask rather than a bounding box for each object. Pinheiro~\etal~\cite{pinheiro2015learning} obtain bounding boxes from instance segmentations by simply drawing the tightest bounding box around each segmentation mask, and show that their system reaches state-of-the-art performance on an object detection benchmark.

The relation between instance segmentation and semantic segmentation is less clear. Intuitively, the two tasks feel very closely related, but it turns out not to be obvious how to apply the network architectures and loss functions that are successful in semantic segmentation to this related instance task. One key factor that complicates the naive application of the popular softmax cross-entropy loss function to instance segmentation, is the fact that an image can contain an arbitrary number of instances and that the labeling is permutation-invariant: it does not matter which specific label an instance gets, as long as it is different from all other instance labels.
One possible solution is to set an upper limit to the number of detectable instances and to impose extra constraints on the labeling, but this may unnecessarily limit the representational power of the network and introduce unwanted biases, leading to unsatisfying results.

Most recent works on instance segmentation with deep networks go a different route. Two popular approaches introduce a multi-stage pipeline with object proposals~\cite{hariharan2014simultaneous, chen2015multi, pinheiro2015learning, pinheiro2016learning, dai2015convolutional, dai2015instance, xu2016gland, arnab2016bottom}, or train a recurrent network end-to-end with a custom loss function that outputs instances sequentially~\cite{park2015learning, romera2015recurrent, ren2016end}. Another line of research is to train a network to transform the image into a representation that is clustered into individual instances with a post-processing step~\cite{silberman2014instance, zhang2015monocular, uhrig2016pixel, liang2015proposal}. Our method belongs to this last category, but takes a more principled (less ad-hoc) approach than previous works and reduces the post-processing step to a minimum.

Inspired by the success of siamese networks~\cite{bromley1993signature, chopra2005learning} and the triplet loss~\cite{weinberger2009distance, schroff2015facenet} in image classification, we introduce a discriminative loss function to replace the pixel-wise softmax loss that is commonly used in semantic segmentation. Our loss function enforces the network to map each pixel in the image to an n-dimensional vector in feature space, such that feature vectors of pixels that belong to the same instance lie close together while feature vectors of pixels that belong to different instances lie far apart. The output of the network can easily be clustered with a fast and simple post-processing operation. With this mechanism, we optimize an objective that avoids the aforementioned problems related to variable number of instances and permutation-invariance.

Our work mainly focuses on the loss function, as we aim to be able to re-use network architectures that were designed for semantic segmentation: we plug in an off-the-shelf architecture 
and retrain the system with our discriminative loss function. 
In our framework, the tasks of semantic and instance segmentation can be treated in a consistent and similar manner and do not require changes on the architecture side. 

The rest of this paper is structured as follows. First we give an extensive overview of the related work in section~\ref{sec:relatedwork}. In section~\ref{sec:method} we discuss our proposed method in detail. In section~\ref{sec:experiments} we set up experiments on two instance segmentation benchmarks and show that we get a performance that is competitive with the state-of-the-art.

\section{Related Work}
\label{sec:relatedwork}
In the last few years, deep networks have achieved impressive results in semantic and instance segmentation. All top-performing methods across different benchmarks use a deep network in their pipeline. 
Here we discuss these prior works and situate our model between them. 

%

\textbf{Proposal-based}
Many instance segmentation approaches build a multi-stage pipeline with a separate object proposal and classification step.
Hariharan~\etal~\cite{hariharan2014simultaneous} and Chen~\etal~\cite{chen2015multi} use MCG~\cite{arbelaez2014multiscale} to generate category-independent region proposals, followed by a classification step.
%
Pinheiro~\etal~\cite{pinheiro2015learning, pinheiro2016learning} use the same general approach, but their work focuses on generating segmentation proposals with a deep network.
Dai~\etal~\cite{dai2015convolutional, dai2015instance} won the 2015 MS-COCO instance segmentation challenge with a cascade of networks (MNC) to merge bounding boxes, segmentation masks and category information. Many works were inspired by this approach and also combine an object detector with a semantic segmentation network to produce instances~\cite{xu2016gland, arnab2016bottom, hayder2016shape}.
In contrast to these works, our method does not rely on object proposals or bounding boxes but treats the image holistically, which we show to be beneficial for handling certain tasks with complex occlusions as discussed in section~\ref{subsec:prosandcons}.
%

\textbf{Recurrent methods}
Other recent works~\cite{park2015learning, romera2015recurrent, ren2016end} employ recurrent networks to generate the individual instances sequentially.
%
Stewart~\etal~\cite{stewart2015end} train a network for end-to-end object detection using an LSTM~\cite{hochreiter1997long}. Their loss function is permutation-invariant as it incorporates the Hungarian algorithm to match candidate hypotheses to ground-truth instances.
%
Inspired by their work, Romera~\etal~\cite{romera2015recurrent} propose an end-to-end recurrent network with convolutional LSTMs that sequentially outputs binary segmentation maps for each instance. 
%
Ren~\etal~\cite{ren2016end} improve upon~\cite{romera2015recurrent} by adding a box network to confine segmentations within a local window and skip connections instead of graphical models to restore the resolution at the output. Their final framework consists of four major components: an external memory and networks for box proposal, segmentation and scoring. 
We argue that our proposed method is conceptually simpler and easier to implement than these methods. Our method does not involve recurrent mechanisms and can work with any off-the-shelf segmentation architecture. 
Moreover, our loss function is permutation-invariant by design, without the need to resort to a Hungarian algorithm.
%

\textbf{Clustering}
Another approach is to transform the image into a representation that is subsequently clustered into discrete instances.
%
Silberman~\etal~\cite{silberman2014instance} produce a segmentation tree and use a coverage loss to cut it into non-overlapping regions.
%
Zhang~\etal~\cite{zhang2015monocular} impose an ordering on the individual instances based on their depth, and use a MRF to merge overlapping predicted patches into a coherent segmentation. Two earlier works \cite{yang2012layered, tighe2014scene} also use depth information to segment instances.
%
Uhrig~\etal~\cite{uhrig2016pixel} train a network to predict each pixel's direction towards its instance center, along with monocular depth and semantic labels. They use template matching and proposal fusion techniques to extract the individual instances from this representation.
%
Liang~\etal~\cite{liang2015proposal} predict pixel-wise feature vectors representing the ground truth bounding box of the instance it belongs to. With the help of a sub-network that predicts an object count, they cluster the output of the network into individual instances.
Our work is similar to these works in that we have a separate clustering step, but our loss does not constrain the output of the network to a specific representation like instance center coordinates or depth ordering; it is less ad-hoc in that sense.
%

\textbf{Other}
Bai~\etal~\cite{bai2016deep} use deep networks to directly learn the energy
of the watershed transform.
A drawback of this bottom-up approach is that they cannot handle occlusions where instances are separated into multiple pieces.   
Kirillov~\etal~\cite{kirillov2016instancecut} use a CRF, but with a novel MultiCut formulation to combine semantic segmentations with edge maps to extract instances as connected regions. A shortcoming of this method is that, although they reason globally about instances, they also cannot handle occlusions. 
Arnab~\etal~\cite{arnab2017pixelwise} combine an object detector with a semantic segmentation module using a CRF model. By considering the image holistically it can handle occlusions and produce more precise segmentations.
%

\textbf{Loss function} 
Our loss function is inspired by earlier works on distance metric learning, discriminative loss functions and siamese networks~\cite{bromley1993signature, chopra2005learning, hadsell2006dimensionality, weinberger2009distance, koestinger2012large}.
Most similar to our loss function, Weinberger~\etal~\cite{weinberger2009distance} propose to learn a distance metric for large margin nearest neighbor classification. Kostinger~\etal~\cite{koestinger2012large} further explore a similar LDA based objective.
More recently Schroff~\etal~\cite{schroff2015facenet}, building on Sun~\etal~\cite{wang2014learning}, introduced the \textit{triplet loss} for face recognition. The triplet loss enforces a margin between each pair of faces from one person, to all other faces. 
Xie~\etal~\cite{xie2015unsupervised} propose a clustering objective for unsupervised learning. 
Whereas these works employ a discriminative loss function to optimize distances between \textit{images} in a dataset, our method operates at the pixel level, optimizing distances between \textit{individual pixels} in an image. To our knowledge, we are the first to successfully use a discriminative loss based on distance metric learning principles for the task of instance segmentation with deep networks.

\section{Method}
\label{sec:method}
\subsection{Discriminative loss function}

\begin{figure}
	\begin{center}
		\includegraphics[width=1.0\linewidth]{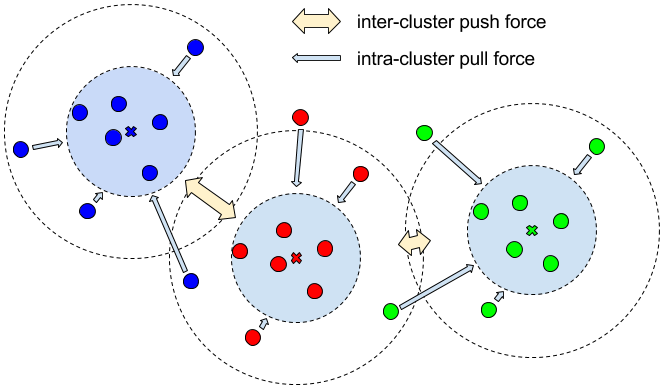}
	\end{center}
	\caption{The intra-cluster pulling force pulls embeddings towards the cluster center, i.e. the mean embedding of that cluster. The inter-cluster repelling force pushes cluster centers away from each other. Both forces are hinged: they are only active up to a certain distance determined by the margins $\delta_v$ and $\delta_d$, denoted by the dotted circles. This diagram is inspired by a similar one in~\cite{weinberger2009distance}.}
	\label{fig:loss}
\end{figure}
Consider a differentiable function that maps each pixel in an input image to a point in n-dimensional feature space, referred to as the \textit{pixel embedding}. The intuition behind our loss function is that embeddings with the same label (same instance) should end up close together, while embeddings with a different label (different instance) should end up far apart.

Weinberger~\etal~\cite{weinberger2009distance} propose a loss function with two competing terms to achieve this objective: a term to penalize large distances between embeddings with the same label, and a term to penalize small distances between embeddings with a different label.

In our loss function we keep the first term, but replace the second term with a more tractable one: instead of directly penalizing small distances between every pair of differently-labeled embeddings, we only penalize small distances between the \textit{mean} embeddings of different labels. 
If the number of different labels is smaller than the number of inputs, this is computationally much cheaper than calculating the distances between every pair of embeddings. This is a valid assumption for instance segmentation, where there are orders of magnitude fewer instances than pixels in an image.

We now formulate our discriminative loss in terms of push (i.e. repelling) and pull forces between and within clusters. A cluster is defined as a group of pixel embeddings sharing the same label, e.g. pixels belonging to the same instance. 
Our loss consists of three terms:

\begin{enumerate}
	\item \textbf{variance term}: an intra-cluster pull-force that draws embeddings towards the mean embedding, i.e.~the cluster center.
	\item \textbf{distance term}: an inter-cluster push-force that pushes clusters away from each other, increasing the distance between the cluster centers.
	\item \textbf{regularization term}: a small pull-force that draws all clusters towards the origin, to keep the activations bounded.
\end{enumerate}

The variance and distance terms are hinged: their forces are only active up to a certain distance. Embeddings within a distance of $\delta_v$ from their cluster centers are no longer attracted to it, which means that they can exist on a local manifold in feature space rather than having to converge to a single point. Analogously, cluster centers further apart than $2 \delta_d$ are no longer repulsed and can move freely in feature space. Hinging the forces relaxes the constraints on the network, giving it more representational power to achieve its goal. 
The interacting forces in feature space are illustrated in figure~\ref{fig:loss}. 

\begin{figure*}
	\begin{center}
		\includegraphics[width=1.0\linewidth]{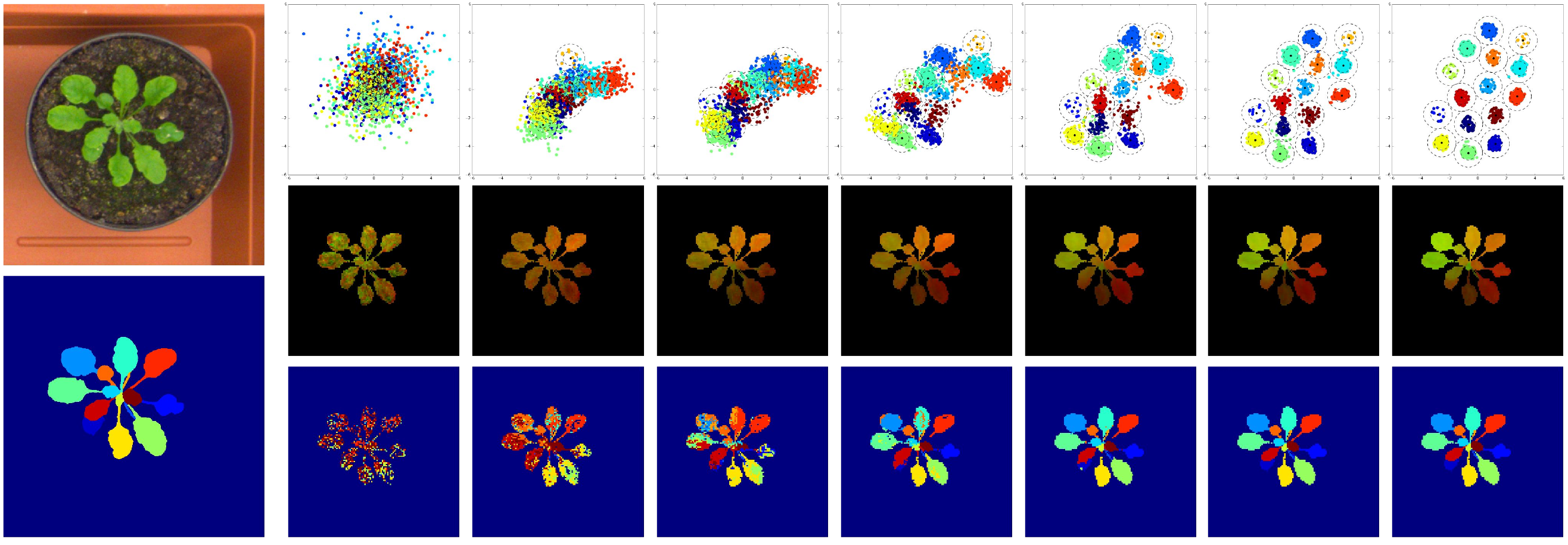}
	\end{center}
	\caption{Convergence of our method on a single image in a 2-dimensional feature space. Left: input and ground truth label. The middle row shows the raw output of the network (as the R- and G- channels of an RGB image), masked with the foreground mask. The upper row shows each of the pixel embeddings $x_i$ in 2-d feature space, colored corresponding to their ground truth label. The cluster center $\mu_c$ and margins $\delta_v$ and $\delta_d$ are also drawn. The last row shows the result of clustering the embeddings by thresholding around their cluster center, as explained in section \ref{sec:postprocessing}. We display the images after 0, 2, 4, 8, 16, 32 and 64 gradient update steps.}
	\label{fig:convergence}
\end{figure*}

The loss function can also be written down exactly. We use the following definitions: $C$ is the number of clusters in the ground truth, $N_c$ is the number of elements in cluster $c$, $x_i$ is an embedding, $\mu_c$ is the mean embedding of cluster $c$ (the cluster center), $\lVert \cdot \rVert$ is the L1 or L2 distance, and $\left[ x \right]_{+} = \textrm{max}(0,x)$ denotes the hinge. $\delta_{\textrm{v}}$ and $\delta_{\textrm{d}}$ are respectively the margins for the variance and distance loss.
The loss can then be written as follows: 

\begin{equation}
	L_{var} = \frac{1}{C} \sum_{c=1}^{C} \frac{1}{N_c} \sum_{i=1}^{N_c} \left[ \lVert \mu_c - x_i \rVert - \delta_{\textrm{v}} \right]_{+}^2
\end{equation}

\begin{equation}
L_{dist} = \frac{1}{C (C-1)} \mathop{\sum_{c_A = 1}^{C} \sum_{c_B = 1}^{C}}_{c_A \neq c_B} \left[ 2 \delta_{\textrm{d}} - \lVert \mu_{c_A} - \mu_{c_B} \rVert \right]_{+}^2
\end{equation}

\begin{equation}
L_{reg} = \frac{1}{C} \sum_{c=1}^{C} \lVert \mu_{c} \rVert
\end{equation}

\begin{equation}
L = \alpha \cdot L_{var} + \beta \cdot L_{dist} + \gamma \cdot L_{reg}
\end{equation}

In our experiments we set $\alpha = \beta = 1$ and $\gamma = 0.001$. The loss is minimized by stochastic gradient descent.

\textbf{Comparison with softmax loss}
We discuss the relation of our loss function with the popular pixel-wise multi-class cross-entropy loss, often referred to as the \textit{softmax loss}. In the case of a softmax loss with $n$ classes, each pixel embedding is driven to a \textit{one-hot vector}, i.e. the unit vector on one of the axes of an $n$-dimensional feature space. Because the softmax function has the normalizing property that its outputs are positive and sum to one, the embeddings are restricted to lie on the unit simplex. When the loss reaches zero, all embeddings lie on one of the unit vectors. By design, the dimensions of the output feature space (which correspond to the number of feature maps in the last layer of the network) must be equal to the number of classes. To add a class after training, the architecture needs to be updated too.

In comparison, our loss function does not drive the embeddings to a specific point in feature space. The network could for example place similar clusters (e.g. two small objects) closer together than dissimilar ones (e.g. a small and a large object). When the loss reaches zero, the system of push and pull forces has minimal energy and the clusters have organized themselves in n-dimensional space. Most importantly, the dimensionality of the feature space is independent of the number of instances that needs to be segmented. Figure~\ref{fig:convergence} depicts the convergence of our loss function when overfitting on a single image with $15$ instances, in a 2-dimensional feature space.

\subsection{Post-processing}
\label{sec:postprocessing}
When the variance and distance terms of the loss are zero, the following is true:

\begin{itemize}
	\item all embeddings are within a distance of $\delta_v$ from their cluster center
	\item all cluster centers are at least $2 \delta_d$ apart
\end{itemize}

If $\delta_d > \delta_v$, then each embedding is closer to its own cluster center than to any other cluster center. It follows that during inference, we can threshold with a bandwidth $b = \delta_v$ around a cluster center to select all embeddings belonging to that cluster. Thresholding in this case means selecting all embeddings $x_i$ that lie within a hypersphere with radius $b$ around the cluster center $x_c$: 
\begin{equation}
x_i \in C \Leftrightarrow \lVert x_i - x_c \rVert < b
\end{equation}
For the tasks of classification and semantic segmentation, with a fixed set of classes, this leads to a simple strategy for post-processing the output of the network into discrete classes: after training, calculate the cluster centers of each class over the entire training set. During inference, threshold around each of the cluster centers to select all pixels belonging to the corresponding semantic class. This requires that the cluster centers of a specific class are the same in each image, which can be accomplished by coupling the cluster centers across a mini-batch.

For the task of instance segmentation things are more complicated. As the labeling is permutation invariant, we cannot simply record cluster centers and threshold around them during inference. We could follow a different strategy: if we set $\delta_d > 2 \delta_v$, then each embedding is closer to all embeddings of its \textit{own} cluster than to any embedding of a \textit{different} cluster. It follows that we can threshold around any embedding to select all embeddings belonging to the same cluster. The procedure during inference is to select an unlabeled pixel, threshold around its embedding to find all pixels belonging to the same instance, and assign them all the same label. Then select another pixel that does not yet belong to an instance and repeat until all pixels are labeled. 

\textbf{Increasing robustness} In a real-world problem the loss on the test set will not be zero, potentially causing our clustering algorithm for instance segmentation to make mistakes. If a cluster is not compact and we accidentally select an outlier to threshold around, it could happen that a real cluster gets predicted as two sub-clusters. To avoid this issue, we make the clustering more robust against outliers by applying a fast variant of the mean-shift algorithm~\cite{FukunagaMeanShift}. As before, we select a random unlabeled pixel and threshold around its embedding. Next however, we calculate the mean of the selected group of embeddings and use the mean to threshold again. We repeat this process until mean convergence.
This has the effect of moving to a high-density area in feature space, likely corresponding to a true cluster center. In the experiments section, we investigate the effect of this clustering algorithm by comparing against \textit{ground truth clustering}, where the thresholding targets are calculated as an average embedding over the ground truth instance labels.

\subsection{Pros and cons}
\label{subsec:prosandcons}
\begin{figure}[t]
	\begin{center}
		\includegraphics[width=1.0\linewidth]{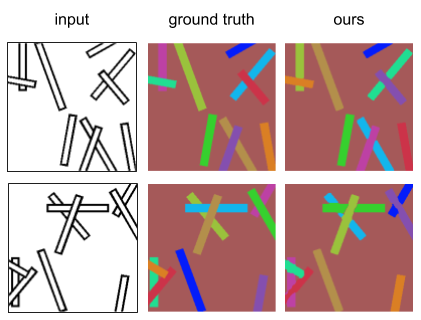}
	\end{center}
	\caption{Results on the synthetic scattered sticks dataset to illustrate that our approach is a good fit for problems with complex occlusions.}
	\label{fig:mikado}
\end{figure}
Our proposed method has some distinctive advantages and disadvantages compared to other methods that we now discuss.

One big limitation of detect-and-segment approaches that is not immediately apparent from their excellent results on popular benchmarks, is that they rely on the assumption that an object's segmentation mask can be unambiguously extracted from its bounding box. 
This is an implicit prior that is very effective for datasets like MS COCO and Pascal VOC, which contain relatively blobby objects that do not occlude each other in complex ways. 
However, the assumption is problematic for tasks where an object's bounding box conveys insufficient information to recover the object's segmentation mask. 
Consider the synthetic \textit{scattered sticks} dataset shown in figure~\ref{fig:mikado} as an example to illustrate the issue. 
When two sticks overlap like two crossed swords, their bounding boxes are highly overlapping. Given only a detection in the form of a bounding box, it is exceedingly hard to unambigously extract a segmentation mask of the indicated object. 
Methods that rely on bounding boxes in their pipeline~\cite{hariharan2014simultaneous, pinheiro2015learning, pinheiro2016learning, dai2015convolutional, dai2015instance, xu2016gland, arnab2016bottom, park2015learning, ren2016end, liang2015proposal, arnab2017pixelwise, hayder2016shape} all suffer from this issue. In contrast, our method can handle such complex occlusions without problems as it treats the image holistically and learns to reason about occlusions, but does not employ a computationally expensive CRF like~\cite{arnab2017pixelwise}.
Many real-world industrial or medical applications (conveyor belt sorting systems, overlapping cell and chromosome segmentation, etc.) exhibit this kind of occlusions. To the best of our knowledge no sufficiently large datasets for such tasks are publicly available, which unfortunately prevents us from showcasing this particular strength of our method to the full.

On the other hand, our method also has some drawbacks. Due to the holistic treatment of the image, our method performs well on 
datasets with a lot of similarity across the images (traffic scenes in Cityscapes or leaf configurations in CVPPP), but
underperforms on datasets where objects can appear in random constellations and diverse settings, like Pascal VOC and MSCOCO. A sliding-window detection-based approach with non-max suppression is more suited for such datasets. For example, if our method were trained on images with only one object, it would perform badly on an image that unexpectedly contained many of these objects. A detection-based approach has no trouble with this. 

\section{Experiments}
\label{sec:experiments}
We test our loss function on two instance segmentation datasets: The CVPPP Leaf Segmentation dataset and the Cityscapes Instance-Level Semantic Labeling Task. These datasets contain a median number of more than 15 instances per image. We also study the influence of the different components of our method and point out where there is room for improvement. 

\subsection{Datasets}
\begin{description}
	\item[CVPPP leaf segmentation]
    The LSC competition of the CVPPP workshop~\cite{CVPPP} is a small but challenging benchmark.
    The task is to individually segment each leaf of a plant. The dataset~\cite{minervini2015finely} was developed to encourage the use of computer vision methods to aid in the study of plant phenotyping~\cite{minervini2015image}. We use the A1 subset which consists of 128 labeled images and 33 test images. \cite{scharr2016leaf} gives an overview of results on this dataset. We compare our performance with some of these works and two other recent approaches~\cite{romera2015recurrent, ren2016end}.
	We report two metrics defined in  ~\cite{scharr2016leaf}: Symmetric Best Dice ($SBD$), which denotes the accuracy of the instance segmentation and Absolute Difference in Count ($|DiC|$) which is the absolute value of the mean of the difference between the predicted number of leaves and the ground truth over all images.
		
	\item[Cityscapes] The large-scale Cityscapes dataset~\cite{cordts2016cityscapes} focuses on semantic understanding of urban street scenes. It has a benchmark for pixel-level and instance-level semantic segmentation. We test our method on the latter, using only the fine-grained annotations. The dataset is split up in 2975 training images, 500 validation images, and 1525 test images. We tune hyperparameters using the validation set and only use the train set to train our final model. We compare our results with the published works in the official leaderboard~\cite{cityscapes}.
    We report accuracy using 4 metrics defined in ~\cite{cordts2016cityscapes}: mean Average Precision (AP), mean Average Precision with overlap of 50\% (AP0.5), AP50m and AP100m, where  evaluation is restricted to objects within 50 m and 100 m distance, respectively.

\end{description}

\subsection{Setup}

\textbf{Model architecture and general setup}
Since we want to stress the fact that our loss can be used with an off-the-shelf network, we use the ResNet-38 network~\cite{wu2016wider}, designed for semantic segmentation. We finetune from a model that was pre-trained on CityScapes semantic segmentation.

In the following experiments, all models are trained using Adam, with a learning rate of 1e-4 on a NVidia Titan X GPU. 

\textbf{Leaf segmentation}
Since this dataset only consists of 128 images, we use online data augmentation to prevent the model from overfitting and to increase the overall robustness. 
We apply random left-right flip, random rotation with $\theta \in \left[0, 2\pi \right]$ and random scale deformation with $s \in \left[1.0, 1.5\right]$. All images are rescaled to 512x512 and concatenated with an x- and y- coordinate map with values between -1 and 1. We train the network with margins $\delta_v = 0.5$, $\delta_d = 1.5$, and 16 output dimensions.
Foreground masks are included with the test set, since this challenge only focuses on instance segmentation.

\textbf{Cityscapes}
Our final model is trained on the training images, downsampled to $768 \times 384$. 
Because of the size and variability of the dataset, there is no need for extra data augmentation.
We train the network with margins $\delta_v = 0.5$, $\delta_d = 1.5$, and 8 output dimensions.
In contrast to the CVPPP dataset, Cityscapes is a \textit{multi-class} instance segmentation challenge. 
Therefore, we run our loss function independently on every semantic class, so that instances belonging to the same class are far apart in feature space, whereas instances from different classes can occupy the same space. For example, the cluster centers of a pedestrian and a car that appear in the same image are not pushed away from each other. We use a pretrained ResNet-38 network~\cite{wu2016wider} to generate segmentation masks for the semantic classes.

\subsection{Analysis of the individual components}
\label{sec:invividualComponentAnalysis}
The final result of the semantic instance segmentation is influenced by three main components: the performance of the network architecture with our loss function, the quality of the semantic labels, and the post-processing step. To disentangle the effects of the semantic segmentation and the post-processing and to point out potential for improvement, we run two extra experiments on the Cityscapes validation set:

\begin{itemize}
	\item \textbf{Semantic segmentation vs ground truth} For the Cityscapes challenge, we rely on semantic segmentation masks to make a distinction between the different classes. Since our instance segmentation will discard regions not indicated in the semantic segmentation labels, the results will be influenced by the quality of the semantic segmentation masks. To measure the size of this influence, we also report performance with the ground truth semantic segmentation masks.
    
    \item \textbf{Mean shift clustering vs ground truth clustering} Since the output of our network needs to be clustered into discrete instances, the clustering method can potentially influence the accuracy of the overall instance segmentation. In this experiment, we measure this influence by clustering with our adapted mean shift clustering algorithm versus thresholding around the mean embeddings over the ground truth instance masks, as explained in section \ref{sec:postprocessing}).

\end{itemize}

\begin{table}
	\begin{center}
		\begin{tabular}{l|c|c}
			& SBD & $|DiC|$ \\
			\hline
            RIS+CRF~\cite{romera2015recurrent} & $66.6$ & $1.1$ \\ 
            MSU~\cite{scharr2016leaf} & $66.7$ & $2.3$ \\ 
			Nottingham~\cite{scharr2016leaf} & $68.3$ & $3.8$ \\ 
			Wageningen~\cite{yin2014multi} & $71.1$ & $2.2$ \\
			IPK~\cite{pape20143} & $74.4$ & $2.6$ \\
            PRIAn~\cite{giuffrida2016learning} & - & $1.3$ \\
            End-to-end~\cite{ren2016end} & $\mathbf{84.9}$ & $\mathbf{0.8}$ \\
            \hline
			Ours & $84.2$ & $1.0$ \\
		\end{tabular}
	\end{center}
	\caption{Segmentation and counting performance on the test set of the CVPPP leaf segmentation challenge.}
	\label{tab:resultsLeafs}
\end{table}

\begin{figure*}
	\begin{center}
        \includegraphics[width=0.95\linewidth]{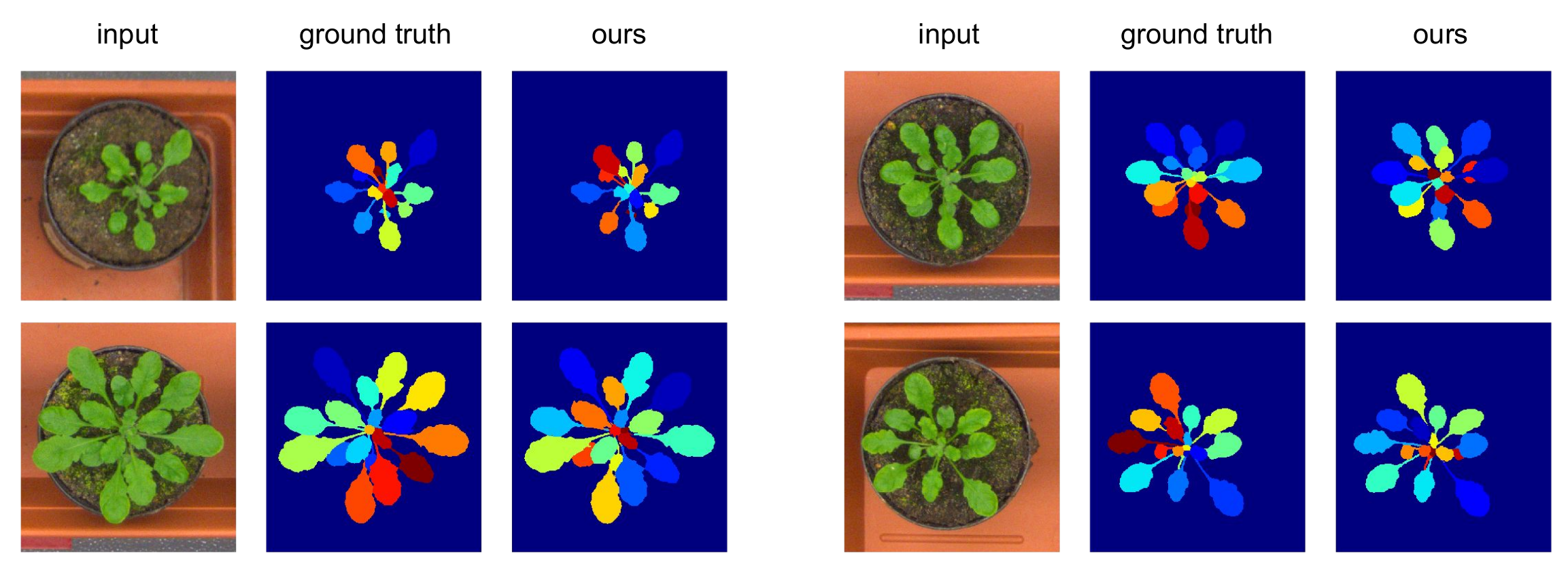}
	\end{center}
	\caption{Some visual examples on the CVPPP leaf dataset.}
	\label{fig:resultsLeafs}
\end{figure*}

\subsection{Results and discussion}

\begin{table}
	\begin{center}
		\begin{tabular}{l|c|c|c|c}
			& AP & AP0.5 & AP100m & AP50m \\
			\hline
			R-CNN+MCG & $4.6$ & $12.9$ & $7.7$ & $10.3$ \\
			FCN+Depth & $8.9$ & $21.1$ & $15.3$ & $16.7$ \\
            JGD & $9.8$ & $23.2$ & $16.8$ & $20.3$ \\
            InstanceCut & $13.0$ & $27.9$ & $22.1$ & $26.1$ \\
            Boundary-aware & $17.4$ & $36.7$ & $29.3$ & $34.0$ \\
            DWT & $19.4$ & $35.3$ & $31.4$ & $36.8$ \\
            Pixelwise DIN & $20.0$ & $38.8$ & $32.6$ & $37.6$ \\
            Mask R-CNN & $26.2$ & $49.9$ & $37.6$ & $40.1$ \\
            \hline
			Ours & $17.5$ & $35.9$ & $27.8$ & $31.0$ \\
		\end{tabular}
	\end{center}
	\caption{Segmentation performance on the test set of the Cityscapes instance segmentation benchmark.}
	\label{tab:resultsCityscapes}
\end{table}

\begin{figure*}
	\begin{center}
		\includegraphics[width=1.0\linewidth]{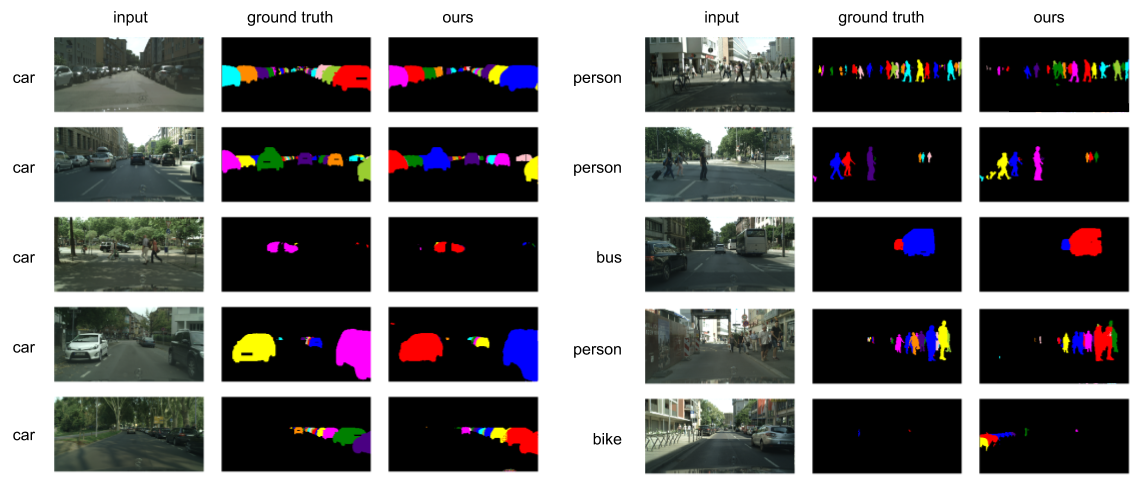}
	\end{center}
	\caption{Some examples for different semantic classes on the Cityscapes instance segmentation validation set. Note some typical failure cases in the last row: incorrect merging of true instances and wrong semantic segmentation masks.}
	\label{fig:resultsCityscapes}
\end{figure*}
            
\begin{table}
	\begin{center}
		\begin{tabular}{l|l||c|c}
			semantic segm. & clustering & AP & AP0.5\\
			\hline
			resnet38~\cite{wu2016wider} & mean-shift & $21.4$ & $40.2$ \\
			resnet38~\cite{wu2016wider} & center threshold & $22.9$ & $44.1$ \\
            ground truth & mean-shift & $37.5$ & $58.5$ \\
            ground truth & center threshold & $47.8$ & $77.8$ \\
		\end{tabular}
	\end{center}
	\caption{Effect of the semantic segmentation and clustering components on the performance of our method on the Cityscapes validation set. We study this by gradually replacing each component with their ground truth counterpart. Row 1 vs row 3: the quality of the semantic segmentation has a big influence on the overall performance. Row 1 vs 2 and row 3 vs 4: the effect of the clustering method is less pronounced but also shows room for improvement.}
	\label{tab:resultsLeafsIndividualComponents}
\end{table}

Figure \ref{fig:resultsLeafs} shows some results of our method on the validation set of the CVPPP dataset. The network makes very few mistakes: only the segmentation of the smallest leafs and the leaf stalks sometimes show a small error. Table~\ref{tab:resultsLeafs} contains the numerical results. We achieve competitive results (SBD score of 84.2) that are on-par with the state-of-the art (SBD score of 84.9). We outperform all non-deep learning methods and also the recurrent instance segmentation of~\cite{romera2015recurrent}, with a method that is arguably less complex.

Figure~\ref{fig:resultsCityscapes} shows some visual results on the Cityscapes validation set. We see that even in difficult scenarios, with street scenes containing a lot of cars or pedestrians, our method often manages to identify the individual objects. Failure cases mostly involve the splitting up of a single object into multiple instances or incorrect merging of neighboring instances. 
This happens in the lower left example, where the two rightmost cars are merged. 
Another failure mode is incorrect semantic segmentation: in the lower right example, the semantic segmentation network accidentally mistakes an empty bicycle storage for actual bikes. The instance segmentation network is left no choice but to give it a shot, and tries to split up the imaginary bikes into individual objects.
Nevertheless, we achieve competitive results on the Cityscapes leaderboard~\cite{cityscapes}, outperforming all but one unpublished work~\cite{arnab2017pixelwise}. 
Note that we perform on-par with the MNC-based method SAIS~\cite{dai2015instance, hayder2016shape} on this dataset. 
See table~\ref{tab:resultsCityscapes} for a complete overview. A video of the results is available at \url{https://youtu.be/FB_SZIKyX50}.

As discussed in section~\ref{sec:invividualComponentAnalysis}, we are interested to know the influence of the quality of the semantic segmentations and the clustering algorithm on the overall performance. The results of these experiments can be found in table~\ref{tab:resultsLeafsIndividualComponents}.
As expected, the performance increases when we switch out a component with its ground truth counterpart. The effect of the semantic segmentation is the largest: comparing the first row (our method) to the third row, we see a large performance increase when replacing the ResNet-38~\cite{wu2016wider} semantic segmentation masks with the ground truth masks. This can be explained by the fact that the average precision metric is an average over the semantic classes. Some classes like tram, train and bus almost never have more than one instance per image, causing the semantic segmentation to have a big influence on this metric. It is clear that the overall performance can be increased by having better semantic segmentations. 

The last two entries of the table show the difference between ground truth clustering and mean shift clustering, both using ground truth segmentation masks. Here also, there is a performance gap. The main reason is that the loss on the validation set is not zero which means the constraints imposed by the loss function are not met. Clustering using mean-shift will therefore not lead to perfect results. The effect is more pronounced for small instances, as also noticeable in the shown examples.

\subsection{Speed-accuracy trade-off}

\begin{table}[t]
	\begin{center}
	\scalebox{0.85}{
		\begin{tabular}{l|c|c|c|c|c|c}
        	 & Dim & AP & AP\textsubscript{gt} & fps & \#p & mem \\
            \hline
             & 512 x 256 & 0.19 & 0.21& 145 & & 1.00 \\
             ENet~\cite{paszke2016enet} & 768 x 384 & 0.21 & 0.25& 94 &  0.36 & 
1.03 \\
             & 1024 x 512 & 0.20 & 0.26& 61 & & 1.12\\
            \hline
             & 512 x 256 & 0.20 & 0.22 & 27 & & 1.22\\
             Segnet~\cite{badrinarayanan2015segnet} & 768 x 384 & 0.22 & 0.26 & 
14 & 29.4 & 1.29 \\
             & 1024 x 512 & 0.18 & 0.24 & 8 & & 1.47\\       
            \hline
             & 512 x 256 & 0.21 & 0.24& 15 & & 2.20\\
             Dilation~\cite{yu2015multi} & 768 x 384 & 0.24 & 0.29 & 6 & 134.3 
& 
2.64 \\
             & 1024 x 512 & 0.23 & 0.30 & 4 & & 3.27\\    
            \hline
             & 512 x 256 & 0.24 & 0.27 &  12 & & 4.45\\
             Resnet38~\cite{wu2016wider} & 768 x 384 & 0.29 & 0.34 & 5 & 124 & 
8.83 \\  
		\end{tabular}
	      }
	\end{center}
	\caption{Average Precision (AP), AP using gt 
segmentation labels (AP\textsubscript{gt}), speed of forward pass (fps), number 
of parameters ($\times10^6$) and memory usage (GB) for different models 
evaluated on the car class of the Cityscapes validation set.}
	\label{tab:resultsExperiments}
\end{table}

To investigate the trade-off between speed, accuracy and memory requirements, 
we train four different network models on different resolutions and evaluate them 
on the car class of the Cityscapes validation set. This also illustrates the 
benefit that our method can be used with any off-the-shelf network designed for 
semantic segmentation.

Table \ref{tab:resultsExperiments} shows the results. We can conclude 
that Resnet-38 is best for accuracy, but requires some more memory. 
If speed is important, ENet would favor over Segnet since it is much faster with 
almost the same accuracy. It also shows that running on a higher resolution 
than 768x384 doesn't increase accuracy much for the tested networks. Note that the post-processing step can be implemented efficiently, causing only a negligible overhead.

\section{Conclusion}
\label{sec:conclusion}
In this work we have proposed a discriminative loss function for the task of instance segmentation. After training, the output of the network can be clustered into discrete instances with a simple post-processing thresholding operation that is tailored to the loss function. Furthermore, we showed that our method can handle complex occlusions as opposed to popular detect-and-segment approaches. Our method achieves competitive performance on two benchmarks. 
In this paper we still used a pretrained network to produce the semantic segmentation masks. 
We will investigate the joint training of instance and semantic segmentation with our loss function in future work.

{\bf Acknowledgement:} This work was supported by Toyota, and was carried out at the TRACE Lab at KU Leuven (Toyota Research on Automated Cars in Europe - Leuven).

{\small
\bibliographystyle{ieee}
\bibliography{ref}
}

\end{document}